\pdfoutput=1
\PassOptionsToPackage{dvipsnames}{xcolor}
\documentclass[sigconf]{acmart}

\usepackage{booktabs} % For formal tables
\usepackage{braket}
\usepackage{xcolor}
\usepackage{tikz}
\usetikzlibrary{pgfplots.groupplots}
\usepackage{pgfplots}
\usepackage{latexsym}
\usepackage{amsfonts}
\usepackage{amsmath}
\usepackage{enumitem}
\usepackage[usestackEOL]{stackengine}
\usepackage{xspace}
 \usepackage{booktabs}
 \usepackage{comment}
\usepackage{hhline}
\usepackage{multirow}
\usepackage[ruled,vlined]{algorithm2e}
\usepackage{graphicx}
\usepackage{verbatim}
\usepackage{algorithm2e}
\usepackage{subcaption}
\usepackage{adjustbox}

\def \dquotes {\let \lq \dlq \let \rq \drq \let \togquotes \squotes }%
\def \framework{FPL\xspace}%
\def \frameworklong{Federated Pair-wise Learning\xspace}%

\copyrightyear{2021} 
\acmYear{2021} 
\setcopyright{acmcopyright}\acmConference[SAC '21]{The 36th ACM/SIGAPP Symposium on Applied Computing}{March 22--26, 2021}{Virtual Event, Republic of Korea}
\acmBooktitle{The 36th ACM/SIGAPP Symposium on Applied Computing (SAC '21), March 22--26, 2021, Virtual Event, Republic of Korea}
\acmPrice{15.00}
\acmDOI{10.1145/3412841.3442010}
\acmISBN{978-1-4503-8104-8/21/03}

\begin{document}
\title[How to Put Users in Control of their Data in Federated  Recommendation]{How to Put Users in Control of their Data in Federated Top-N Recommendation with Learning to Rank}

\author{Vito Walter Anelli, Yashar Deldjoo, Tommaso Di Noia, Antonio Ferrara, Fedelucio Narducci}
\authornote{Authors in alphabetical order. Corresp.: Antonio Ferrara (antonio.ferrara@poliba.it).}

\affiliation{%
  \institution{Politecnico di Bari, Bari, Italy}
}
\email{firstname.lastname@poliba.it}

\renewcommand{\shortauthors}{V. W. Anelli et al.}

\begin{abstract}
Recommendation services are extensively adopted in several user-centered applications as a tool to alleviate the information overload problem and help users in orienteering in a vast space of possible choices. In such scenarios, data ownership is a crucial concern since users may not be willing to share their \textit{sensitive} preferences (e.g., visited locations) with a central server.
Unfortunately, data harvesting and collection is at the basis of modern, state-of-the-art approaches to recommendation. 
To address this issue, we present \framework, an architecture in which users collaborate in training a central factorization model while controlling the amount of sensitive data leaving their devices. 
The proposed approach implements pair-wise learning-to-rank optimization by following the \textit{Federated Learning} principles, originally conceived to mitigate the privacy risks of traditional machine learning. The public implementation is available at \url{https://split.to/sisinflab-fpl}.

\end{abstract}




\maketitle

\section{Introduction}
\label{sec:introduction}
Collaborative filtering (CF) models have been mainstream research in the recommender system (RS) community over the last two decades thanks to their performance accuracy~\cite{DBLP:journals/corr/BokdeGM15a}. 
Among them, a prominent class uses the matrix factorization (MF) approach as the inference model. The MF model's main aim is to uncover user and item latent representations whose linear interaction explains observed feedback.
To date, the majority of existing MF models are trained in a \textit{centralized} fashion causing several concerns about the privacy of user data and discouraging their collection.
The consequent \dquotes{data scarcity} dilemma can thereby jeopardize the training of MF models.
Training high-quality MF models relies on sufficient in-domain interaction data to ensure that enough co-occurrence information exists to shape similar behavioral/preference patterns in a user community. 
In recent years, federated learning (FL) was proposed by Google as a mean to offer a \textit{privacy-by-design} solution~\cite{anelli2019towards,DBLP:conf/aistats/McMahanMRHA17} for machine-learned models. Federated learning aims to meet ML privacy shortcomings by horizontally distributing the model's training over user devices; thus, clients exploit private data without sharing them~\cite{DBLP:conf/aistats/McMahanMRHA17}.
Weiss\textit{ et al.}~\cite{DBLP:conf/ifip9-6/Weiss07} state that privacy can be preserved by limiting data collection, which is one of the main privacy concerns~\cite{DBLP:series/ccn/JeckmansBEHLT13}. The accuracy of RS based on the CF paradigm is dependent on the amount of user preferences available.
Our idea is to put users in control of their sensitive data by allowing them to choose the amount of information to share with the server. Hence, if data collection from the server side is reduced, other threats related to retention, sales, and unauthorized data browsing are limited. The proposed system, \framework (short for \frameworklong), is a \textit{federated} factorization model for collaborative recommendation.
It extends state-of-the-art factorization approaches to build a RS that puts users in control of their sensitive data. Users participating in the federation process can decide \textit{if} and \textit{to which extent} they are willing to disclose their \textit{sensitive} private data (i.e., what they liked/consumed). 
\framework mainly leverages not-sensitive information (e.g., places the user has not visited) -- which can be large and non-sensitive -- to reach a competitive accuracy and, at the same time, respect a satisfactory balance between accuracy and privacy.
We have carried out extensive experiments on real-world datasets~\cite{DBLP:journals/tist/YangZQ16} in the Point of Interest (PoI) domain by considering the accuracy of recommendation and diversity metrics. 
The experimental evaluation shows that \framework can provide high-quality recommendations, putting the user in control of the amount of sensitive data to share.

\section{Background}
\label{sec:bkg_tech}
\paragraph{Federated Learning.} 
Federated learning (FL) is a paradigm initially envisioned by Google~\cite{DBLP:journals/corr/KonecnyMRR16, DBLP:conf/aistats/McMahanMRHA17} to train a machine-learning model from data distributed among a loose federation of users' devices (e.g., personal mobile phones).
The rationale is to face the increasing issues of ownership and locality of data to mitigate the privacy risks resulting from centralized machine learning~\cite{kairouz2019advances} while improving personalization~\cite{DBLP:conf/bdc/JalaliradSCS19}.
In particular, given $\Theta$ denoting the parameters of a machine learning model, we consider a learning scenario where the objective is to minimize a generic loss function $G(\Theta)$.
FL is a learning paradigm in which the users $u \in \mathcal{U}$ of a federation collaborate to solve the learning problem under the coordination of a central server $S$ without sharing or exchanging their raw data with $S$.
From an algorithmic point of view, we start with $S$ sharing $\Theta$ with the federation of devices.
Then, specific methods solve a local optimization problem on the single device. 
The client shares the parameters of its local model with $S$. The parameters provided by the clients are then used to update $\Theta$, which is sent back to the devices in a new iteration step.

\vspace{-0.5em}
\paragraph{Factorization Models and Pair-Wise Recommendation} 
\label{sec:MF}
A recommendation problem over a set of users $\mathcal{U}$ and a set of items $\mathcal{I}$ is defined as the activity of finding for each user $u \in \mathcal{U}$ an item $i \in \mathcal{I}$ that maximizes a utility function $g : \mathcal{U} \times \mathcal{I} \rightarrow \mathbb{R}$.
In this context, $\mathbf{X} \in\mathbb{R}^{|\mathcal{U}| \times |\mathcal{I}|}$ is the user-item matrix containing for each $x_{ui}$ an explicit or implicit feedback (e.g., rating or check-in, respectively) of user $u \in \mathcal{U}$ for item $i \in \mathcal{I}$.
In the work at hand, an implicit feedback scenario is considered --- i.e., feedback is, e.g., purchases, visits, clicks, views, check-ins ---, with $\mathbf{X}$ containing binary values. Therefore, $x_{ui} = 1$ and $x_{ui} = 0$ denote either user $u$ has consumed or not item $i$, respectively. 
In \framework, the underlying data model is a Factorization model, inspired by MF~\cite{DBLP:journals/computer/KorenBV09}, a recommendation model that became popular in the last decade thanks to its state-of-the-art recommendation accuracy~\cite{DBLP:journals/corr/BokdeGM15a}. 
This technique aims to build a model $\Theta$ in which each user $u$ and each item $i$ is represented by the embedding vectors $\mathbf{p}_u$ and $\mathbf{q}_i$, respectively, in the shared latent space $\mathbb{R}^F$. The algorithm relies on the assumption that $\mathbf{X}$ can be factorized such that the dot product between $\mathbf{p}_u$ and $\mathbf{q}_i$ can explain any observed user-item interaction $x_{ui}$, and that any non-observed interaction can be estimated as
$\hat{x}_{ui}(\Theta) = b_i(\Theta) + \mathbf{p}_u^T(\Theta)\cdot \mathbf{q}_i(\Theta) $
where $b_i$ is a term denoting the bias of the item $i$.
Among pair-wise approaches for learning-to-rank the items of a
catalog, Bayesian Personalized Ranking (BPR)~\cite{DBLP:conf/uai/RendleFGS09} is one of the most broadly adopted, thanks to its capabilities to correctly rank with \textit{acceptable} computational complexity.
In detail, given a training set defined by $\mathcal{K}=\{(u,i,j) \;|\; x_{ui} = 1 \land x_{uj} = 0 \}$, 
 BPR solves the optimization problem via the criterion $\underset{\Theta}{\max} \sum_{(u,i,j) \in \mathcal{K}} \ln \ \sigma (\hat{x}_{uij}(\Theta)) - \lambda  \lVert \Theta \rVert^2$,
where $\hat{x}_{uij}(\Theta) = \hat{x}_{ui}(\Theta) - \hat{x}_{uj}(\Theta)$ is a real value modeling the relation between user $u$, item $i$ and item $j$, $\sigma$ is the sigmoid function, and $\lambda$ is a regularization parameter to prevent overfitting.
Pair-wise optimization can be applied to a wide range of recommendation models, included factorization. 
Hereafter, we denote the  model $\Theta = \Braket{\mathbf{P}, \mathbf{Q}, \mathbf{b}}$, where $\mathbf{P} \in \mathbb{R}^{|\mathcal{U}|\times F}$ is a matrix whose $u$-th row corresponds to the vector $\mathbf{p}_u$, and $\mathbf{Q} \in \mathbb{R}^{|\mathcal{I}|\times F}$ is a matrix in which the $i$-th row corresponds to the vector $\mathbf{q}_i$. Finally, $\mathbf{b} \in \mathbb{R}^{|\mathcal{I}|}$ is a vector whose $i$-th element corresponds to the value $b_i$.

\vspace{-0.5em}
\section{Approach}
\label{sec:protocol}
Following the federated learning principles, let $\mathcal{U}$ be the set of users (clients) with a server $S$ coordinating them. 
Assume users consume items from a catalog $\mathcal{I}$ and give feedback about them (as in the recommendation problem of Section~\ref{sec:MF}).
$S$ is aware of the catalog $\mathcal{I}$, while exclusively user $u$ knows her own set of consumed items.

To setup the federation for \framework, a global model is built on $S$ such that $\Theta_S = \Braket{\mathbf{Q}, \mathbf{b}}$, where $\mathbf{Q} \in \mathbb{R}^{|\mathcal{I}|\times F}$ and $\mathbf{b} \in \mathbb{R}^{|\mathcal{I}|}$ are the item-factor matrix and the bias vector (introduced in Section~\ref{sec:bkg_tech}).
Conversely, on each user $u$'s device \framework builds a model $\Theta_u = \Braket{\mathbf{p}_u}$, which corresponds to the representation of user $u$ in the latent space of dimensionality $F$. Hence, $\Theta_u$ and $\Theta_S$ are privately combined together. The client produces tailored recommendations by scalar multiplying local $\mathbf{p}_u$ and $\mathbf{q}_i$.
Each user $u$ holds her own private dataset $\mathbf{x}_u \in \mathbb{R}^{\mathcal{I}}$, which, analogously to a centralized recommender system, corresponds to the $u$-th row of matrix $\mathbf{X}$.
Each \framework client $u$ hosts a user-specific training set $\mathcal{K}_u : \mathcal{U} \times \mathcal{I} \times \mathcal{I}$ defined by $\mathcal{K}_u =\{(u,i,j) \;|\; x_{ui}=1 \land x_{uj}=0 \}$, where $x_{ui}$ represents the $i$-th element of $x_u$.
Please note that we refer to $X^+ = \sum_{u \in \mathcal{U}}|\{x_{ui}\;|\;x_{ui}=1\}|$ as the number of positive interactions.

The classic BPR-MF learning procedure~\cite{DBLP:conf/uai/RendleFGS09} for model training can not be applied to the federated learning scheme~\cite{DBLP:conf/aistats/McMahanMRHA17}.
Instead, we propose a novel learning paradigm that
%is executed for a number $E$ of epochs and
works by rounds of communication
%that
and envisages \textbf{Distribution} $\mathbf{\rightarrow}$ \textbf{Computation} $\mathbf{\rightarrow}$ \textbf{Transmission} $\mathbf{\rightarrow}$ \textbf{Aggregation} sequences between the server and the clients, whose details are as follows.

\noindent (1) \textbf{Distribution.} $S$ randomly selects a subset of users $\mathcal{U}^- \subseteq \mathcal{U}$ and delivers them the model $\Theta_S$. \\
\noindent (2) \textbf{Computation.} Each user $u$ generates $T$ triples $(u, i, j)$ from her dataset $\mathcal{K}_u$ and for each of them performs BPR stochastic optimization to compute the updates for the local $\mathbf{p}_u$ vector of $\Theta_u$, and for $\mathbf{q}_i$, $b_i$, $\mathbf{q}_j$, and $b_j$ of the received $\Theta_S$, following:
\begin{gather*}
\label{eq:update}
\Delta\theta = \frac{e^{-\hat{x}_{uij}}}{1+e^{-\hat{x}_{uij}}} \cdot \frac{\partial}{\partial \theta}\hat{x}_{uij} - \lambda \theta, \\
\text{with} \quad
    \frac{\partial}{\partial \theta}\hat{x}_{uij} =
\begin{cases} 
(\mathbf{q}_i - \mathbf{q}_j) & \mbox{if } \theta=\mathbf{p}_u, \\
\mathbf{p}_u & \mbox{if } \theta=\mathbf{q}_i,\qquad -\mathbf{p}_u \quad \mbox{if } \theta=\mathbf{q}_j, \\
1 & \mbox{if } \theta=b_i,\qquad -1\hphantom{1} \quad \mbox{if } \theta=b_j.\\
\end{cases}
\end{gather*}
It is worth noticing that Rendle~\cite{DBLP:conf/uai/RendleFGS09} suggests, in a centralized scenario, to adopt a uniform distribution (over $\mathcal{K}$) to choose the training triples randomly. The purpose is to avoid data traversed item-wise or user-wise, since this may lead to slow convergence. Conversely, in a federated approach, we required to train the model user-wise since the training of each round of communication is performed separately on each client $u$ knowing only data in $\mathcal{K}_u$. This is the reason why, in \framework, the designer can control of the number of triples $T$ used for training, to tune the degree of local computation --- i.e., how much the sampling is user-wise traversing. \\
\noindent (3) \textbf{Transmission.} The clients in $\mathcal{U}^-$ send back to $S$ a portion of the updates ($\Delta\Theta_{S,u}$) for the computed item factor vector and item bias.
More in detail, since the training output of a triple $(u, i, j)$ in BPR lets the server distinguish the consumed item $i$ from the non-consumed one $j$ (for example just by analyzing the positive and the negative sign of $\Delta b_i$ and $\Delta b_j$), while they show the same absolute value, we argue that sending all the updates computed by $u$ may allow  $S$ to reconstruct $\mathcal{K}_u$ thus raising a privacy issue.
Since our primary goal is to put users in control of their data, \framework proposes a solution to overcome this vulnerability. 
By sending the sole update  $(\Delta \mathbf{q}_j, \Delta b_j)$ of each training triple $(u, i, j)$, user $u$ would share with $S$ indistinguishably negative or missing values, which are assumed to be \textit{non-sensitive} data. Furthermore, in \framework we introduce the parameter $\pi$, which allows users to control of the number of consumed items to share with the central server $S$. In detail, $\pi$ works as a probability that the update $\Delta\Theta_{S,u}$ contains a specific positive item update $(\Delta\mathbf{q}_i,\Delta b_i)$ in addition to $(\Delta\mathbf{q}_j,\Delta b_j)$. \\
\noindent (4) \textbf{Global aggregation.} $S$ aggregates the received updates in  $\mathbf{Q}$ and  $\mathbf{b}$ to build the new model $\Theta_S \leftarrow \Theta_S + \alpha \sum_{u \in \mathcal{U}^-}
\Delta\Theta_{S,u}$, with $\alpha$ being the learning rate (each row of $\mathbf{Q}$ and each element of $\mathbf{b}$ are updated by summing up the contribution of all clients in $\mathcal{U}^-$ for the corresponding item).

\vspace{-0.4em}
\section{Experiments}
\label{sec:experiments}
{\setlength{\tabcolsep}{0.24em}
\renewcommand{\arraystretch}{0.8}
\begin{table}[t]
  \caption{Characteristics of the datasets used for experiments:
  $\left| \mathcal{U} \right|$, $\left| \mathcal{I} \right|$, and $X^+$ are the number of users, items, and records.}
  %$\left| \mathcal{U} \right|$ is the number of users, $\left| \mathcal{I} \right|$ the number of items, $X^+$ the number of records.} 
  \vspace{-1em}
  \centering
  \label{tbl:dataset_char}
  \small
  \begin{tabular}{c@{\hskip 1.5em}cccccc@{\hskip 1.5em}ccccccccc}
    \toprule
    \textbf{Dataset} & $\left| \mathcal{U} \right|$ & $\left| \mathcal{I} \right|$ & $X^+$ & $\frac{X^+}{\left|\mathcal{U} \right|}$ & $\frac{X^+}{\left| \mathcal{I} \right|}$ & $\frac{X^+}{\left| \mathcal{L} \right| \cdot \left| \mathcal{U} \right|} \%$ \\
    \toprule
    \textbf{Brazil} & 17,473  & 47,270 & 599,958 & 34.34 & 12.69 & 0.00073\%
\\ 
    \textbf{Canada} & 1,340 & 29,518 & 63,514 & 47.40 & 2.15 & 0.00161\% \\
    \textbf{Italy} & 1,353 & 25,522 & 54,088 & 39.98 & 2.20 & 0.00157\%  \\
    \bottomrule 
  \end{tabular}%
        \vspace{-1.5em}
\end{table}
}

\paragraph{Experimental Setting}
\framework needs to be evaluated in a domain that guarantees the availability of transaction data the user may prefer to protect.
In our view, the optimal domain would be that of the Point-of-Interest (PoI), which concerns data that users usually perceive as sensitive.
Among the many available datasets, a good candidate is the \textit{Foursquare} dataset~\cite{DBLP:journals/tist/YangZQ16}, which is often considered as a reference for evaluating PoI recommendation models.
To mimic a federation of devices in a single country, we have extracted check-ins for three countries, namely Brazil, Canada, and Italy to obtain datasets with different size/sparsity characteristics.
To evaluate \framework, we have kept users with more than $20$ interactions to avoid the known CF cold-start limitations.
We have split the datasets by adopting a realistic temporal hold-out 80-20 splitting on a per-user basis~\cite{DBLP:reference/sp/GunawardanaS15} (see training sets' characteristics in Table~\ref{tbl:dataset_char}).

To evaluate the efficacy of \framework, we have conducted the experiments by considering non-personalized methods (random and most popular recommendation), and different recommendation approaches, including the centralized \textbf{BPR-MF} implementation~\cite{DBLP:conf/uai/RendleFGS09}, \textbf{VAE}~\cite{liang2018variational}, and  \textbf{FCF}~\cite{DBLP:journals/corr/abs-1901-09888}, which is, to date, the only federated recommendation approach based on MF (since no source code is available, we reimplemented and considered it in the reader's interest). 
To evaluate the impact of feedback deprivation on recommendation accuracy, we have evaluated different values of $\pi$ in the range $[0.0$, $1.0]$,
with $\pi = 0.0$ meaning that $u$ is not sharing any positive feedback with $S$, and $\pi = 1.0$ meaning that $u$ is sharing the updates on all positive items.
Hence, we have considered two different configurations regarding computation and communication:

\begin{itemize}[leftmargin=*]
    \item \textbf{\textit{s}FPL}: it reproduces the centralized stochastic learning, where the central model is updated sequentially; thus, we set $|\mathcal{U}^-|=1$ to involve just one random client per round, and it extracts solely one triple $(u,i,j)$ from its dataset ($T=1$) for the training phase;
    \item \textbf{\textit{p}FPL}: we enable parallelism by involving all clients in each round of communication ($\mathcal{U^-}=\mathcal{U}$); we keep $T=1$.
\end{itemize}

\noindent In Rendle\textit{ et al.}~\cite{DBLP:conf/uai/RendleFGS09}, authors suggest to set the number of triples in one epoch of BPR to $X^+$, which corresponds to the number of optimizations steps. A particular choice is to randomly sampling $T = \frac{X^+}{|\mathcal{U}|}$ triples per user.
To compare federated training with BPR and among configurations, we consider $rpe$ rounds of communication of \framework to be equivalent to one epoch of centralized BPR, if $rpe$ is set such that we perform the same overall number of optimization steps.
This results in $rpe = X^+$ for \textit{s}FPL, and  $rpe = \frac{X^+}{|\mathcal{U}^-|}$ for \textit{p}FPL.

\vspace{-0.5em}
\paragraph{Reproducibility}
For the splitting strategy, we have adopted a \textbf{temporal hold-out $80$/$20$} to separate our datasets in training and test set. Moreover, to find the most promising learning rate $\alpha$, we have further split the training set, adopting a temporal hold-out $80$/$20$ strategy on a user basis to extract her validation set.
\textbf{VAE} has been trained by considering three autoencoder topologies, with the following number of neurons per layer: $200$-$100$-$200$, $300$-$100$-$300$, $600$-$200$-$600$. We have chosen candidate models by considering the best models after training for $50$, $100$, and $200$ epochs, respectively. For the \textbf{factorization models}, we have performed a grid search in BPR-MF for $\alpha \in \{0.005, 0.05, 0.5\}$ varying the number of latent factors in $\{10, 20, 50\}$. Then, to ensure a fair comparison, we have exploited the same learning rate and number of latent factors to train \framework and \textbf{FCF}, and we explored the models in the range of $\{10,\ldots, 50\}$ iterations. We have set \textit{user-} and \textit{positive item-}regularization parameter to $\frac{1}{20}$ of the learning rate.
The \textit{negative item-}regularization parameter is $\frac{1}{200}$ of the learning rate, as suggested by Anelli\textit{ et al.}~\cite{DBLP:conf/recsys/AnelliNSPR19}.

\vspace{-0.5em}
\paragraph{Evaluation Metrics}\label{sec:metrics}
We have evaluated the performance of \framework under the accuracy and diversity perspective.
The accuracy of the models is measured by exploiting Precision ($P@N$) and Recall ($R@N$). They respectively represent, for each user, the proportion of relevant recommended items in the recommendation list, and the fraction of relevant items that have been altogether suggested. We have assessed the statistical significance of results by adopting Student's paired T-test considering p-values $<0.05$ (see complete results at \url{https://split.to/sisinflab-fpl}). The results are in general statistically significant but the differences among BPR-MF, \textit{s}FPL, and \textit{p}FPL, which is a very important result.
To measure the diversity of recommendations, we have measured the Item Coverage ($IC@N$), and the Gini Index ($G@N$).
$IC$ provides the normalized number of diverse items recommended to users. It also conveys the sense of the degree of personalization \cite{DBLP:journals/tkde/AdomaviciusK12}.
Gini measures distributional inequality, i.e., how unequally different items a RS provides users with~\cite{DBLP:reference/sp/CastellsHV15}. A higher value of $G$ corresponds to higher personalization~\cite{DBLP:reference/sp/GunawardanaS15}.

\vspace{-0.5em}
\paragraph{Discussion}
\begin{table*}[tbp]
\begin{adjustbox}{valign=t,minipage={.7\textwidth}}
\small
\centering
    \captionof{table}{Results of accuracy and beyond-accuracy metrics for baselines and \framework on the three datasets. For each configuration of \framework, the experiment with the best $\pi$ is shown. For all metrics, the greater the better.}
      \vspace{-1em}
{
\renewcommand{\arraystretch}{0.82}
\setlength{\tabcolsep}{-1.5pt}
\small
\begin{tabular}{l@{\extracolsep{5pt}}cccc@{\extracolsep{5pt}}cccc@{\extracolsep{5pt}}cccc}
\toprule
 & \multicolumn{ 4}{c}{\textbf{Brazil}} & \multicolumn{ 4}{c}{\textbf{Canada}} & \multicolumn{ 4}{c}{\textbf{Italy}} \\ \cline{2-5} \cline{6-9} \cline{10-13}
 & \textbf{P@10} & \textbf{R@10} & \textbf{IC@10} & \textbf{G@10} & \textbf{P@10} & \textbf{R@10} & \textbf{IC@10} & \textbf{G@10} & \textbf{P@10} & \textbf{R@10} & \textbf{IC@10} & \textbf{G@10} \\ \toprule
\textbf{Random}  & 0.00013	& 0.00015	& 0.97567 &	0.70946  & 0.00030 & 0.00035 & 0.36639 & 0.26809  & 0.00030 & 0.00029 & 0.41055 &	0.28914
%\textbf{Random}  & 0.00013	& 0.00015	& 46120 &	0.70946  & 0.00030 & 0.00035 & 10815 & 0.26809  & 0.00030 & 0.00029 & 10478 &	0.28914
 \\ \hline
\textbf{Top-Pop}  & 0.01909 &	0.02375	& 0.00040	& 0.00020
 & 0.04239 &	0.04679	& 0.00061	& 0.00030 &	0.04634	& 0.05506 &	0.00074 &	0.00035 \\ \hline
%\textbf{Top-Pop}  & 0.01909 &	0.02375	& 19	& 0.00020 & 0.04239 &	0.04679	& 18	& 0.00030 &	0.04634	& 0.05506 &	19 &	0.00035 \\ \hline
\textbf{VAE} *   & 0.10320 &	0.13153	& 0.11642	& 0.02117	& 	0.06060	& 0.06317& 	0.03537 &	0.00652	 &	0.01459 & 0.02985 &	0.00647	& 0.00327
%\textbf{VAE} *   & 0.10320 &	0.13153	& 5503	& 0.02117	& 	0.06060	& 0.06317& 	1044 &	0.00652	 &	0.01459 & 0.02985 &	165	& 0.00327
\\ \hline
\textbf{BPR-MF}   & 0.07702 & 0.09494 & 0.05399 & 0.00756  & 0.03694 & 0.03650 & 0.04120 & 0.00998 & 0.04560 & 0.05458 & 0.00074 & 0.00036 \\ \hline
%\textbf{BPR-MF}   & 0.07702 & 0.09494 & 2552 & 0.00756  & 0.03694 & 0.03650 & 1216 & 0.00998 & 0.04560 & 0.05458 & 19 & 0.00036 \\ \hline
\textbf{FCF}  & 0.03089	& 0.03749 &	0.01927	& 0.00095 & 0.03724 &	0.03836	& 0.01707 &	0.00174	&	0.03126	& 0.03708	& 0.01579	& 0.00158 \\ \hline
%\textbf{FCF}  & 0.03089	& 0.03749 &	911	& 0.00095 & 0.03724 &	0.03836	& 504 &	0.00174	&	0.03126	& 0.03708	& 403	& 0.00158 \\ \hline
\textbf{\textit{s}FPL} ** & 0.07757 & 0.09581 & 0.03345 & 0.00561 &  0.04515 & 0.04550 & 0.01528 & 0.00243 & 0.04701 & 0.05600 & 0.00071 & 0.00036 \\ 
%\textbf{\textit{s}FPL} ** & 0.07757 & 0.09581 & 1581 & 0.00561 &  0.04515 & 0.04550 & 451 & 0.00243 & 0.04701 & 0.05600 & 18 & 0.00036 \\ 
\textbf{\textit{p}FPL} ** & 0.07771 & 0.09582 & 0.04472 & 0.00638 &  0.04582 & 0.04637 & 0.01440 & 0.00213 &  0.04642 & 0.05465 & 0.00376 & 0.00056 \\ \midrule
% \textbf{\textit{p}FPL} ** & 0.07771 & 0.09582 & 2114 & 0.00638 &  0.04582 & 0.04637 & 425 & 0.00213 &  0.04642 & 0.05465 & 96 & 0.00056 \\ \midrule
\multicolumn{13}{l}{* For Italy, VAE does not produce recommendations for all the users; thus, we followed the weighting }\\
\multicolumn{13}{l}{ scheme proposed in prior literature~\cite{DBLP:conf/recsys/MesasB17}}\\
\multicolumn{13}{l}{** Best $\pi$ obtained for FPL variants (Br, Ca, It) are: \textit{s}FPL = (0.5, 0.1, 0.4), \textit{p}FPL = (0.8, 0.1, 1)}\\

\bottomrule
\end{tabular}
}
\label{tbl:general-results}
\end{adjustbox}
\hfill
\begin{adjustbox}{valign=t,minipage={.28\textwidth}}
\small
\begin{tikzpicture}
\begin{groupplot}[
    group style={
        group name=my plots,
        group size=1 by 3,
        xlabels at=edge bottom,
        xticklabels at=edge bottom,
        vertical sep=0pt
    },
    width=\textwidth,
    height=2.7cm,
    xlabel=$\pi$,
    xtick={0.1,0.5,1},
    xlabel near ticks,
    ylabel style={at={(-0.005,0.5)}},
    ylabel style={align=center},
    tickpos=left,
    ytick align=outside,
    xtick align=outside,
        yticklabel style={
    /pgf/number format/fixed,
    /pgf/number format/precision=3
    },
    scaled y ticks=false
]
\nextgroupplot[ylabel={Brazil\\F1}]
    \addplot [MidnightBlue, mark=*] table [x=pi, y=a, col sep=tab] {tables/data/f1_plot_brazil.tsv};
    \addplot [Turquoise, mark=*] table [x=pi, y=c, col sep=tab] {tables/data/f1_plot_brazil.tsv};
\nextgroupplot[ylabel={Canada\\F1}]
    \addplot [MidnightBlue, mark=*] table [x=pi, y=a, col sep=tab] {tables/data/f1_plot_canada.tsv};
    \addplot [Turquoise, mark=*] table [x=pi, y=c, col sep=tab] {tables/data/f1_plot_canada.tsv};
\nextgroupplot[ylabel={Italy\\F1}]
    \addplot [MidnightBlue, mark=*] table [x=pi, y=a, col sep=tab] {tables/data/f1_plot_italy.tsv};
    \addplot [Turquoise, mark=*]table [x=pi, y=c, col sep=tab] {tables/data/f1_plot_italy.tsv};
\end{groupplot}
\end{tikzpicture}
\vspace{-2.6em}

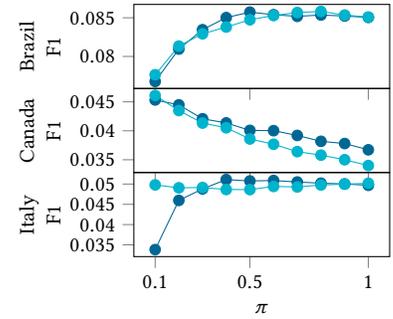
\captionof{figure}{F1 performance at different values of $\pi$ in the range $[0.1,1]$. Dark blue is \textit{s}FPL, light blue \textit{p}FPL.}
\label{fig:varyingpi}
\end{adjustbox}

\vspace{-1.5em}
\end{table*}
The goal of the experiments is assessing whether it is possible to obtain a recommendation performance comparable to a centralized pair-wise learning approach while allowing the users to control their data. 
In this respect, Table \ref{tbl:general-results} shows the accuracy and diversity results of the comparison between the state-of-the-art baselines and the experimental configurations of \framework presented in Section~\ref{sec:experiments}.
By focusing on accuracy metrics, we may notice that VAE outperforms the other approaches in the three datasets.
However, who is familiar with VAE knows that, since it restricts training data by applying k-core, it does not always produce recommendations for all the users.
Moreover, it is important to investigate the differences of \framework with respect to BPR-MF, which is a pair-wise centralized approach, being FPL the first federated pair-wise recommender based on a factorization model.
The performance of BPR-MF against \framework, in the configuration \textit{s}FPL, shows how Precision and Recall in \textit{s}FPL are slightly outperforming BPR-MF while achieving very similar diversity values. 
The consideration that the performance is comparable is surprising since the two methods share the sequential training, but \textit{s}FPL exploits a $\pi$ reduced to $0.5$, $0.1$, and $0.4$, respectively, for Brazil, Canada, and Italy. 
This behavior is more evident in Figure~\ref{fig:varyingpi}, where the harmonic mean between Precision and Recall (F1) is plotted for different values of $\pi$. If we look at the dark blue line, we may observe how the best result does not correspond to $\pi = 1$.
When comparing \textit{p}FPL with \textit{s}FPL, we observe that the increased parallelism does not affect the performance significantly.
As a concluding remark, we may affirm that \dquotes{\textit{the proposed system can generate recommendations with a quality that is comparable with the centralized pair-wise learning approach}}, and that \dquotes{\textit{the training parallelism does not significantly affect results.}}
Compared to FCF, \framework generally behaves better while preserving privacy to a greater extent, since sharing gradients of all rated items in FCF may leak raw data~\cite{DBLP:journals/corr/abs-1906-05108}.
Afterwards, we varied $\pi$ in the range $[0.1, \dots, 1.0]$ to investigate how removal of the updates for consumed items affects the final recommendation accuracy, and we plotted the accuracy performance by considering F1 in Figure~\ref{fig:varyingpi}. The best performance rarely corresponds to $\pi=1$. 
On the contrary, the training reaches a peak for a certain value of $\pi$, and then the system performance decays in accuracy when increasing the amount of shared positive updates.
In rare cases, e.g., \textit{s}FPL, and \textit{p}FPL for Brazil dataset, the decay is absent, but results that are very close for different values of $\pi$. The general behavior suggests that the system learning exploits the updates of positive items to absorb information about popularity. 
This consideration is coherent with the mathematical formulation of the learning procedure, and it is also supported by the observation that for Canada and Italy \framework reaches the peak before with respect to Brazil.
Indeed, Canada and Italy datasets are less sparse than Brazil, and the increase of information about positive items may lead to push up too much the popular items (this is a characteristic of pair-wise learning), while the same behavior in Brazil can be observed for $\pi \simeq 1$. 
Ultimately, \textit{users can receive high-quality recommendations, also when disclosing a small amount of sensitive data}.

\vspace{-0.4em}
\section{Conclusion and Future Work}
\label{sec:conclusions}
We proposed \framework, a novel federated learning framework that exploits pair-wise learning for factorization models.
We have designed a model that leaves the user-specific information of the original factorization model in the clients' devices so that a user may be completely in control of her sensitive data and could share no positive feedback with the server.
The framework can be envisioned as a general factorization model in which clients can tune the amount of information shared among devices.
We have conducted an exploratory, but extensive, experimental evaluation to analyze the degree of accuracy, the diversity of the recommendation results, the trade-off between accuracy, and amount of shared transactions.
We have assessed that the proposed model shows performance comparable with several state-of-the-art baselines and the classic centralized factorization model with pair-wise learning.
The evaluation shows that clients may share a small portion of their data with the server and still receive high-performance recommendations.
We believe that the proposed privacy-oriented paradigm may open the doors to a new class of ubiquitous recommendation engines.

\vspace{-0.3em}
\bibliographystyle{abbrv}
\bibliography{bibliography} 

\end{document}